\documentclass[conference]{IEEEtran}
\IEEEoverridecommandlockouts
\usepackage{cite}
\usepackage{amsmath,amssymb,amsfonts}
\usepackage{graphicx}
\usepackage{textcomp}
\usepackage{xcolor}
\usepackage{txfonts}
\usepackage{booktabs} 

\usepackage{hyperref} 
\hypersetup{hidelinks}

\usepackage{algorithm}
\usepackage{algorithmic}
\usepackage[algo2e]{algorithm2e} 

\usepackage{amsmath} 
\usepackage{amssymb}

\usepackage{threeparttable}
\usepackage{float}

\def\BibTeX{{\rm B\kern-.05em{\sc i\kern-.025em b}\kern-.08em
    T\kern-.1667em\lower.7ex\hbox{E}\kern-.125emX}}

\DeclareUnicodeCharacter{2212}{-}

\begin{document}

\title{SA-CNN: Application to text categorization issues using simulated annealing-based convolutional neural network optimization}

\author{\IEEEauthorblockN{1\textsuperscript{st} Zihao GUO†}
\IEEEauthorblockA{\textit{École Centrale de Nantes} \\
Nantes, France \\
zihao.guo@eleves.ec-nantes.fr}
\and

\IEEEauthorblockN{2\textsuperscript{nd} Yueying CAO}
\IEEEauthorblockA{\textit{École Centrale de Nantes} \\
Nantes, France \\
yueying.cao@eleves.ec-nantes.fr}
}

\maketitle
    \begin{abstract}
Convolutional neural networks (CNNs) are a representative class of deep learning algorithms including convolutional computation that perform translation-invariant classification of input data based on their hierarchical architecture. However, classical convolutional neural network learning methods use the steepest descent algorithm for training, and the learning performance is greatly influenced by the initial weight settings of the convolutional and fully connected layers, requiring re-tuning to achieve better performance under different model structures and data. Combining the strengths of the simulated annealing algorithm in global search, we propose applying it to the hyperparameter search process in order to increase the effectiveness of convolutional neural networks (CNNs). In this paper, we introduce SA-CNN neural networks for text classification tasks based on Text-CNN neural networks and implement the simulated annealing algorithm for hyperparameter search. Experiments demonstrate that we can achieve greater classification accuracy than earlier models with manual tuning, and the improvement in time and space for exploration relative to human tuning is substantial.
\end{abstract}

    \section{Introduction}
In recent years, significant breakthroughs have been achieved in the field of convolutional neural networks for text classification, and Yoon Kim\cite{kim-2014-convolutional} proposed a straightforward single-layer CNN architecture that can outperform traditional algorithms in a variety of uses. Rie Johnson and Tong Zhang\cite{johnson2014effective} applied CNNs to high-dimensional text data and learned with embed small text regions for classification. Tong He and Weilin Huang\cite{he2016text} proposed a convolutional neural network that extracts regions and features related to text from image components. This type of model use vectors to characterize each sentence in the text, which are then merged into a matrix and utilized as input for constructing a CNN network model.

Numerous experiments have shown, however, that the performance of neural networks is highly dependent on their architecture\cite{szegedy2015going}\cite{he2016deep}\cite{krizhevsky2012imagenet}. Due to the discrete nature of these parameters, accurate optimization algorithms cannot be used to resolve the architecture optimization problem\cite{junior2019particle}. Manually tuning the parameters of a model to optimize its performance for different tasks is not only inefficient, but also likely to miss the optimal parameters, resulting in a network architecture that does not achieve maximum performance, which is not advantageous in comparison to traditional classification algorithms\cite{Deng2019application}. In addition, the widely utilized grid search is an enumerative search, i.e., it tries every possibility by performing a cyclic traversal of all candidate parameter choices, which is marked by its high time consumption and limited globalization. Therefore, it is practical to use an algorithm to automatically and fairly rapidly determine the optimal architecture of a neural network.

It has been shown that tuning neural network hyperparameters with metaheuristic algorithms not only simplifies the network\cite{ahmed2019novel}\cite{loussaief2018convolutional}, but also enhances its classification performance. In this paper, we use simulated annealing algorithm to optimize the neural network architecture, and we model the neural network hyperparameter optimization problem as a dual-criteria optimization problem of classification accuracy and computational complexity. The resulting network achieves improved classification performance in the text classification task.
    \section{Background and Related Work}

\subsection{The current utilisation text classification in neural networks}
Text classification was initially done by using knowledge engineering to build an expert system and perform classification, which is a laborious task with limited accuracy. After that, along with the development of statistical learning methods and machine learning disciplines, the classical approach of feature engineering plus shallow classification models developed gradually (Fig.1). During this period, rule-based models: decision trees\cite{su2006fast}, probability-based models: Nave Bayes classification algorithms\cite{mccallum1998comparison}\cite{chen2009feature}, geometry-based models: SVM\cite{wang2006optimal}and statistical models: KNN\cite{yong2009improved}, etc. However, these models typically rely heavily on time-consuming feature engineering or large quantities of additional linguistic resources, and are ineffective at learning semantic information about words.

\begin{figure}[htbp]
\centerline{\includegraphics[width=\columnwidth]{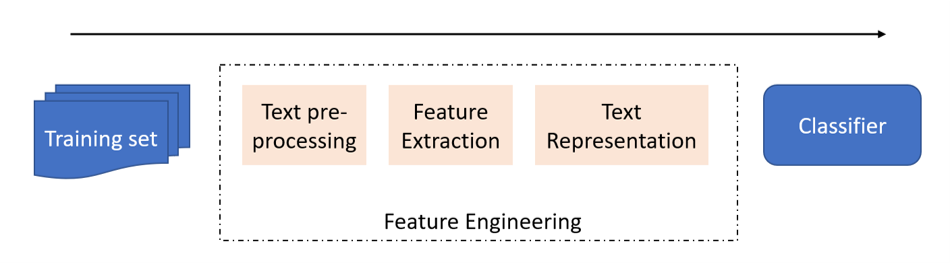}}
\caption{A diagrammatic representation of the process of shallow learning}
\label{fig}
\end{figure}

In recent years, research on deep learning that incorporates feature engineering into the process of model fitting has significantly enhanced the performance of text classification tasks. Kim \cite{kim-2014-convolutional} explored the use of convolutional neural networks with multiple windows for text classification, a method that has been widely adopted in industry due to its high computational speed and parallelizability. Yang et al \cite{yang2016hierarchical} proposed HAN, a hierarchical attention mechanism network that mitigates the gradient disappearance problem caused by RNNs in neural networks. Johnson and Zhang \cite{johnson2017deep} proposed a word-level deep CNN model that improves network performance by increasing network depth without significantly increasing computational burden.

Additionally to convolutional neural networks and recurrent neural networks, numerous researchers have proposed more intricate models in recent years.  The capsule networks-based text classification model proposed by Zhao et al \cite{zhao2018investigating}. outperformed conventional neural networks. Google proposed the BERT model \cite{devlin2018bert}, which overcomes the problem that static word vectors cannot solve the problem of a word having multiple meanings. However, the parameters of each of the aforementioned deep learning models have a substantial effect on network performance and must be optimized for optimal network performance.

\subsection{Current research status on the simulated annealing method and associated neural network optimization}
The Simulated Annealing Technique is a stochastic optimization algorithm based on the Monte Carlo iterative solution approach that was first designed for combinatorial optimization and then adapted for general optimization. Its fundamental concept is based on the significance sampling approach published by Metropolis in 1953, but Kirkpatrick et al.\cite{kirkpatrick1983optimization} did not properly implement it into the field of combinatorial optimization until 1983.

\begin{figure}[htbp]
\centerline{\includegraphics[width=\columnwidth]{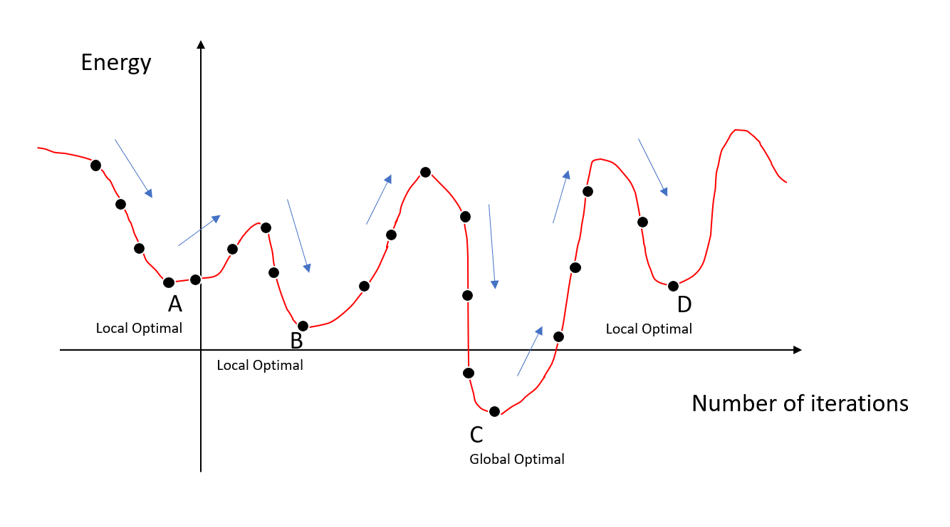}}
\caption{Simulated annealing method picture schematic.}
\label{fig}
\end{figure}

The algorithm for simulated annealing is taken from the process of metal annealing\cite{yong2021new}, which may be summarized roughly as follows. The simulated annealing technique begins with a high initial temperature, then as the temperature parameter decreases, it searches for the optimal solution among all conceivable possibilities. The SA method has a probability of accepting a solution that is worse than the initial one, i.e. the locally optimal solution can probabilistically jump out and finally converge to the global optimum. This likelihood of accepting a suboptimal answer decreases until the SA approaches the global optimal solution.

The standard optimization process of the simulated annealing algorithm can be described as follows.
\begin{algorithm}[H] \SetKwData{Left}{left}\SetKwData{This}{this}\SetKwData{Up}{up} \SetKwFunction{Union}{Union}\SetKwFunction{FindCompress}{FindCompress} \SetKwInOut{Input}{input}\SetKwInOut{Output}{output}
 

\Input{Initial feasible solution: $x_{0}$ ;\\Initial temperature: $T_{0}$ ; \\Termination temperature  : $T_{f}$ ;\\Iteration number: k}

\caption{Simulate Anneal Arithmetic}
$\quad x \longleftarrow x_{k}$, $T \longleftarrow T_{0}$, $k \longleftarrow 0$

\label{alg3} 
\begin{algorithmic}
    
    \WHILE {$k \leqslant k_{max}$ \textbf{and} $T \geqslant T_{f}$}
    \STATE
            $x_{k}\longleftarrow$ NEIGHBOR(s)
            \STATE
            $\delta f=f(x_{k}) - f(x)$
            
    \IF {$\Delta f < 0$ \textbf{or} RANDOM(0, 1) $\leqslant P(\Delta f, T)$}
    \STATE
        $x \longleftarrow x_{k}$

    \ENDIF
    \STATE
            $T \longleftarrow COOLING(T,k,k_{max})$
    \STATE      
            $k \longleftarrow k+1$
         
    \ENDWHILE   

\end{algorithmic} 
\end{algorithm}


SA has been utilized extensively in VLSI\cite{wong2012simulated}, production scheduling\cite{van1992job}, machine learning\cite{andrieu2003introduction}, signal processing\cite{chen1999adaptive}, and other domains as a general-purpose stochastic search method. Boltzmann machine\cite{hinton2007boltzmann}, the first neural network capable of learning internal expressions and solving complicated combinatorial optimization problems, utilizes the SA principle for optimization precisely, therefore the optimization potential of SA is evident.

RasdiRere et al.\cite{rere2015simulated} employed simulated annealing to automatically construct neural networks and alter hyperparameters, and experimental findings revealed that the method could increase the performance of the original CNN, demonstrating the efficacy of this optimization technique. Mousavi et al.\cite{mousavi2017next} updated a solar radiation prediction model by integrating artificial neural networks and simulated annealing with temperature cycling to increase ANN calibration performance. Choubin et al.\cite{choubin2020spatial} utilized the simulated annealing (SA) feature selection approach to find the influential factors of PM modeling based on current air detection machine learning models for the spatial risk assessment of PM10 in Barcelona.

According to the study, however, there is no research on the combination of simulated annealing and neural networks for text categorization tasks. This research conducts tests on neural networks employing simulated annealing to enable automated hyperparameter search, based on the fact that neural networks now generate superior outcomes in text categorization.
    \section{METHODS}
\subsection{Convolutional neural networks for text processing tasks (Text-CNN)}
Convolutional neural networks (CNN) originated in the field of computer vision; however, with the recent deformation of the CNN input layer, this neural network structure has been steadily transferred to the field of natural language processing, where it is often referred to as Text CNN. The schematic is seen below.

\begin{figure}[htbp]
    \centerline{\includegraphics[width=\columnwidth]{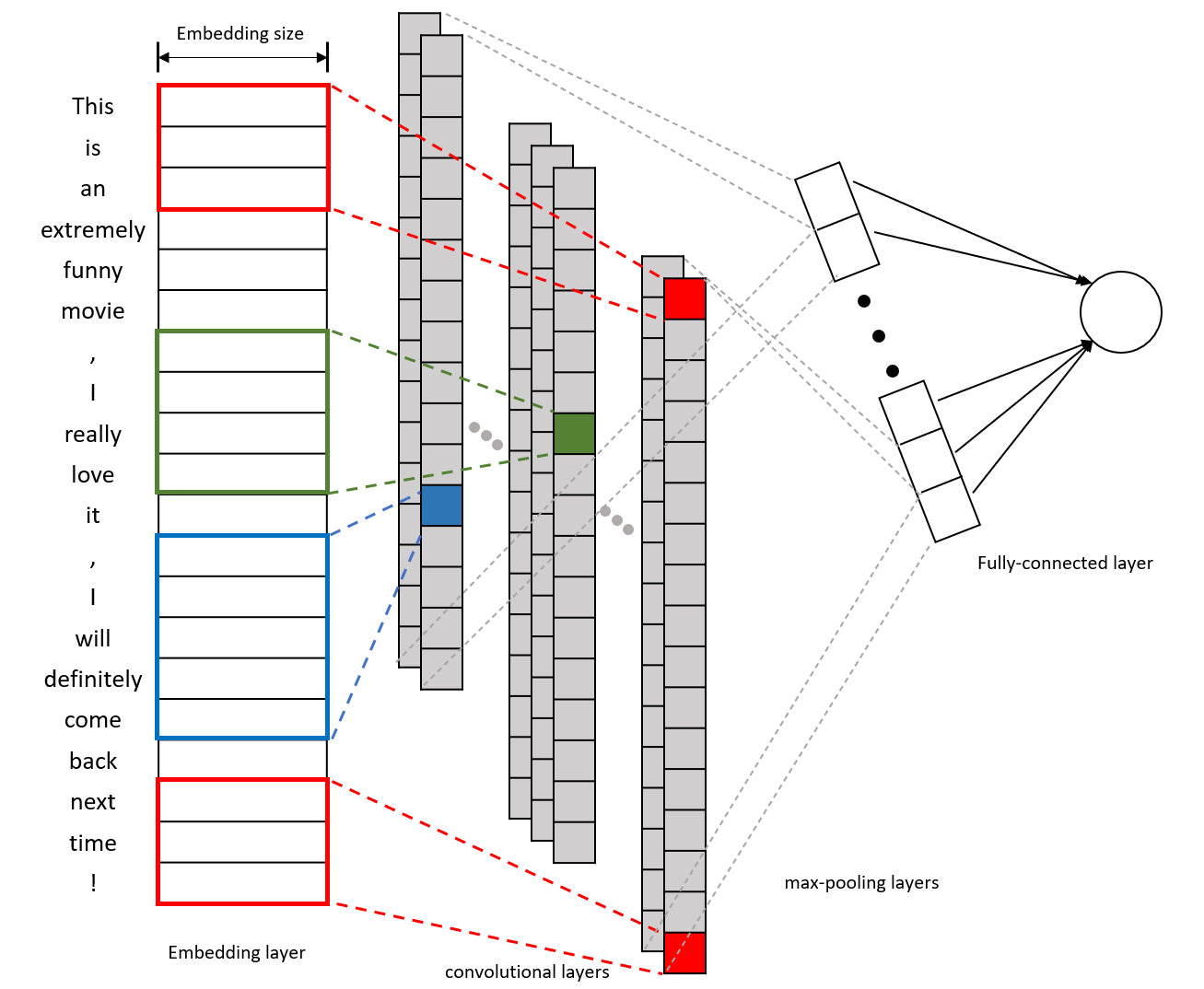}}
    \caption{Schematic diagram of Text-CNN.}
    \label{fig}
\end{figure}

A text statement consists of n words, therefore the text is separated according to words and each word is mapped to the Word Embedding layer as a k-dimensional vector. At this point, for this input model, the text may be regarded as a $n\times k$ single-channel picture. During the processing of the convolution layer, the convolution is used to extract the relationships between tuples containing different numbers of words, i.e., to generate new features and obtain different feature maps, when the width of the convolution kernel is equal to the dimension k of the word vector.

The feature map in the form of an n-dimensional vector is then sampled using max-pooling to determine the maximum value, and the data is pooled and utilized as input to the fully connected layer. The softmax function[Eq. (1)] is then employed to convert these probabilities into discrete 0 or 1 class labels in order to solve this classification challenge.

\begin{equation}
    y = softmax(W_{1}h + b_{1})
\end{equation}

As demonstrated in Eq. (2), a cross-entropy loss function is frequently utilized for the classification job in model training.
\begin{equation}
    Loss = -\sum_{i=1}^{n}y_{i}\times log(y_{i}^{'})
\end{equation}
where $y_{i}$ is the label value corresponding to the real probability distribution of the $i^{th}$ sample, $y_{i}^{'}$ is the prevalence measure corresponding to the projected probability distribution of the $i^{th}$ sample, and n is the number of samples in the training set.

The hyperparameter optimization process of the neural network by simulated annealing method is shown below.
\begin{figure}[htbp]
\centerline{\includegraphics[width=\columnwidth]{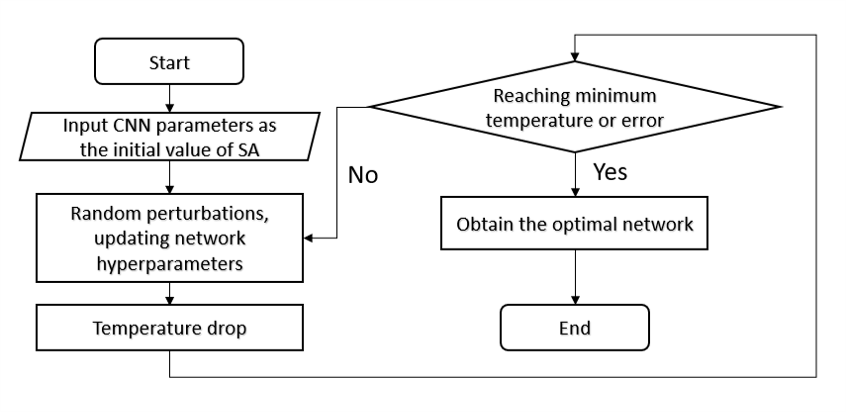}}
\caption{Flowchart of neural network hyperparameter tuning using simulated annealing.}
\label{fig}
\end{figure}

\subsection{Hyperparametric optimization method based on multi-objective simulated annealing method}
In this study, we use the MOSA algorithm proposed by Gülcü and Kuş \cite{gulcu2021multi} to optimize the hyperparameters of the convolutional neural network in order to find the most suitable parameters quickly and achieve a higher accuracy rate. Similar to the single-objective SA algorithm, we extend the SA algorithm, which only considers the error rate of the network implementation, to consider the two objectives of the number of FLOPs required by the network and the error rate of the network, respectively, and define the stopping criterion of the simulated annealing method as the number of iterations.

\subsubsection{Main flow of MOSA algorithm}

The MOSA algorithm primarily uses Smith's Pareto dominance rule \cite{smith2008dominance} due to complications such as the need for two target values to be on the same scale, followed by the application of decision rules to aggregate the probabilities, and the need to maintain different temperatures due to the different probabilities evaluated for each target. All non-dominated solutions encountered during the search are stored in an external archive, A, when the first iteration begins at the initial temperature. As new solutions are accepted, A is updated (by inserting the new solution X' and removing all solutions dominated by it) and a superior solution is eventually formed as the Pareto frontier. As depicted in the following flowchart, whenever a new solution X' is obtained, X and A are updated based on the dominance relationship between the current solution X, the new solution X', and the solution in the external archive A. This process of re-visiting previously visited archive solutions is known as the return-to-base strategy. In contrast to the single-target SA algorithm, the $\Delta F$ calculation used to determine the probability of acceptance is different in this method. For calculation purposes, a single temperature will be maintained regardless of the number of targets.

\begin{figure}[htbp]
\centerline{\includegraphics[width=\columnwidth]{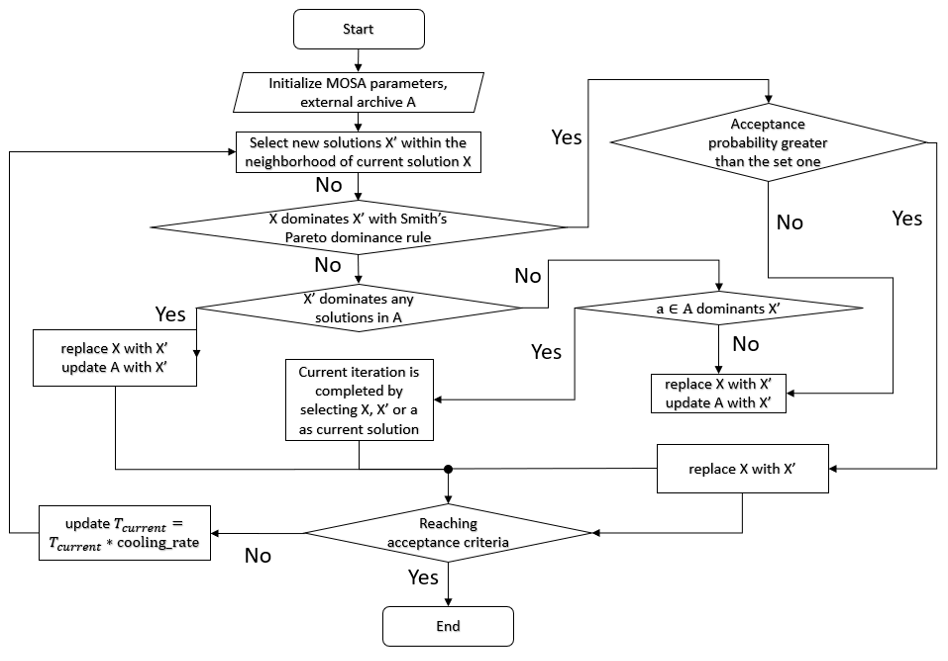}}
\caption{Schematic diagram of MOSA algorithm flow.}
\label{fig}
\end{figure}

\subsubsection{Setting and calculation of temperature and probability for SA method}

First, the initial and finale temperatures. Theoretically, the higher the initial temperature setting, the better, but since the time required for convergence increases correspondingly with the temperature, it is necessary to make a compromise between convergence time and convergence accuracy. To define $T_init$, we apply a real-time initial temperature selection strategy. In this strategy, rather than using Eq. (3) to calculate the initial probability value for a given initial temperature (where $\Delta F$ is the amount of deterioration in the objective function and $T_cur$ is the current temperature). 
\begin{equation}
    p_{acc}=min\{1, exp( −\Delta F/T_{cur} )\}
\end{equation}
We use Eq. (4) to calculate the initial temperature with an initial probability value (where Fave is the average amount of deterioration penalty calculated during short-term "aging"). tfinal is also defined by this method of real-time temperature adjustment)
\begin{equation}
    T_{init}= −(\Delta F_{ave}/(ln(p_{acc})))
\end{equation}

\subsubsection{Acceptance criteria in searching parameters}
In this experiment, the number of iterations is used to define the stopping criterion for the simulated annealing method. The rationale is as follows: if poor early performance is observed on the validation set, the current training process is terminated and the next search round is initiated. This approach has the benefit of introducing noise while decreasing the total running time of the HPO method. In each iteration of the simulated annealing method, the newly generated configuration is trained on the training set of the original training set and evaluated on the validation set (i.e., the test split of the original training set).
We apply the Xavier weight initial value setting term, a learning rate of 0.001, the Rmsprop optimizer, and a batch size of 32.

\subsubsection{Optimization of the Simulated Annealing method}
Starting with the initial solution I and initial temperature $T_{init}$, the iterative process “generate a new solution $\to$ calculate the objective function difference $\to$ accept or reject" is repeated for the current solution and temperature. $T_{cur}$ is gradually decayed, and if the optimal error rate achieved on the validation set does not improve after three consecutive calendar hours, the training procedure is terminated, indicating that the current solution is the approximate optimal solution, i.e. the optimal state of the network has been reached.
    \section{EXPERIMENTS AND RESULTS}

\begin{table*}[htbp]
\caption{Relationship between the number of external and internal iterations calculated for different values of cooling rate.}
\begin{center}
\begin{tabular*}{0.6\linewidth}{@{}llllll@{}}
\toprule
\hline
\textbf{\textit{Iteration budget}}& \textbf{\textit{$T_{init}$}}& \textbf{\textit{$T_{final}$}}& \textbf{\textit{Cooling rate}}& \textbf{\textit{\#Outer iterations}}& \textbf{\textit{\#Inner iterations}}\\

\hline
\midrule
250 & 0.577 & 0.12 & 0.99 & 156.2 & 1.6 \\
250 & 0.577 & 0.12 & 0.95 & 30.6 & 8.1 \\
250 & 0.577 & 0.12 & 0.9 & 14.9 & 16.7 \\
250 & 0.577 & 0.12 & 0.85 & 9.6 & 25.8 \\
250 & 0.577 & 0.12 & 0.8 & 7.0 & 35.5 \\
\bottomrule
\hline
\end{tabular*}
\begin{threeparttable}
    \begin{tablenotes}
        \footnotesize
        \item[*] The table is cited from Gülcü and Kuş's experiment\cite{gulcu2021multi} on the effect of MOSA cooling rate on the number of iterations
    \end{tablenotes}
\end{threeparttable}
\label{tab1}
\end{center}
\end{table*}

\subsection{Introduction to the experimental data set}\label{AA}
This experiment utilized two short text categorization datasets, the MR dataset and the TREC dataset. MR is a dataset for the sentiment analysis of movie reviews, with each sample classified as positive sentiment or negative sentiment. The CR dataset consists of reviews of five electronic products, and these sentences have been manually tagged with the sentiment of the reviews. TREC is a question classification dataset in which each data point represents a question description, and the job entails categorizing a question into six question kinds, including person, place, numerical information, abbreviation, description, and entity.

The table displays the statistical information of the three data sets.
\begin{table}[H]
\caption{Statistics for the text classification dataset}
\begin{center}
\begin{tabular}{ c c c c c c c}
\toprule
\hline
\textbf{\textit{Dataset}}& \textbf{\textit{\#C}}& \textbf{\textit{AvgLen}}& \textbf{\textit{Dsize}}& \textbf{\textit{$|V|$}}& \textbf{\textit{$|V_{pre}|$}}& \textbf{\textit{Test}}\\
\hline
\midrule
MR & 2 & 20 & 10662 & 18765 & 16448 & CV \\
CR & 2 & 19 & 3775 & 5340 & 5046 & CV \\
TREC & 6 & 10 & 5952 & 9592 & 9125 & 50\\
\bottomrule
\hline
\end{tabular}
\begin{threeparttable}
    \begin{tablenotes}
        \footnotesize
        \item[*] C: number of target categories, AvgLen: average sentence length, DSize: dataset size, $|V|$: number of words, $|V_{pre}|$: number in pre-trained word vector, Test: test set size (CV means there was no standard train/test split and thus 10-fold CV was used)
    \end{tablenotes}
\end{threeparttable}
\label{tab1}
\end{center}
\end{table}

Several samples from each of the two datasets are provided below.
\begin{itemize}
\item It's a square, sentimental drama that satisfies, as comfort food often can. [\textbf{MR Dataset}, tags: positive].
\item The sort of movie that gives tastelessness a bad rap. [\textbf{MR Dataset}, tags: negative].
\item this camera is so easy to use ! [\textbf{CR Dataset}, tags: positive].
\item the sound level is also not as high as i would have expected . [\textbf{CR Dataset}, tags: negative].
\item What is Australia's national flower? [\textbf{TREC Dataset}, tags: place]
\item Who was the first man to fly across the Pacific Ocean? [\textbf{TREC Dataset}, tags: person]
\end{itemize}

\subsection{Introduction to the comparison model}
Comparing the model in the article to the experimental model in Kim's\cite{kim-2014-convolutional} study for experimentation.
\begin{itemize}
\item \textbf{CNN-rand: }All word vectors are initialized at random before being utilized as optimization parameters during training.
\item \textbf{CNN-static: }All word vectors are directly acquired with the Word2Vec tool and are fixed.
\item \textbf{CNN-multichannel: }A mix of CNN-static and CNN-non-static, i.e. two types of inputs.
\item \textbf{DCNN\cite{kalchbrenner2014convolutional}: }Dynamic Convolutional Neural Network with k-max pooling.
\item \textbf{MV-RNN\cite{socher2012semantic}: }Matrix-Vector Recursive Neural Network with parse trees.
\end{itemize}

\subsection{SA-CNN parameter setting}

\subsubsection{Parameter setting of MOSA}
In this research, we employed the identical parameter settings for the simulated annealing approach as Gülcü and Kuş\cite{gulcu2021multi}, set the starting initial probability value to 0.5, and derived $T_{init} \approx $ 0.577 and $T_{finial} \approx $ 0.12 in a similar fashion. The link between the number of exterior and internal iterations estimated for different cooling rate values is depicted in the table.2.  \\The number of outer iterations defines the number of search networks, whereas the number of inner iterations determines the number of single network training iterations. As the cooling rate, we chose 0.95, where the number of external cycles is greater than the number of internal cycles, to ensure that as many network structures as possible are searched for and to avoid repeated training of a single network structure as much as possible, thereby avoiding becoming trapped in a local optimal solution of network selection.
\subsubsection{Search range of neural network hyperparameters}
In this study, we utilize a 300-dimensional word2vec trained by Mikolov\cite{mikolov2013distributed} as the initial representation and continually update the word vector during the training process, similar to Kim's\cite{kim-2014-convolutional} approach. In this paper, the empirical range of hyperparameters to be tuned in the network is provided so that the simulated annealing technique can search for a new solution from the existing one. Expanding the range of searchable hyperparameters may result in improved experimental outcomes, provided computational resources permit.

\begin{itemize} 
\item Conv:
    \begin{itemize} 
    \item kernelCount: [32, 64, 96, 100, 128, 160, 256]
    \item dropoutRate: [0.1, 0.2, 0.3, 0.4, 0.5]
    \end{itemize}
\item fullyConnected:
    \begin{itemize} 
    \item unitCount: [16, 32, 64, 128, 256, 512]
    \item dropoutRate: [0.1, 0.2, 0.3, 0.4, 0.5]
    \end{itemize}
\item learningProcess:
    \begin{itemize} 
    \item activation: [relu, leaky\_relu, elu, tanh, linear]
    \item learningRate: [0.0001, 0.001, 0.01, 0.0002, 0.0005, 0.0008, 0.002, 0.004, 0.005, 0.008]
    \item batchSize: [64, 128, 256]
    \end{itemize}
\item seedNumber: 40
\item ratioInit: 0.9
\end{itemize}

\subsection{Experimental results and discussion}

\subsubsection{Comparison of model accuracy results}
The following table shows the accuracy of different CNN models for text classification tasks on MR, CR and TREC datasets.

\begin{table}[H]
\caption{Accuracy of different models on the dataset.}
\begin{center}
\begin{tabular*}{0.7\linewidth}{c c c c}
\toprule
\hline
\textbf{Model}& \textbf{MR}& \textbf{CR}& \textbf{TREC}\\
\hline
\midrule
CNN-rand & 76.1 & 79.8 & 91.2  \\
CNN-static & 81.0 & 84.7 & 92.8  \\
CNN-non-static & \textbf{81.5} & 84.3 & 93.6  \\
CNN-multichannel & 81.1 & \textbf{85.0} & 92.2  \\
D-CNN &  -  & -  & 93.0 \\
MV-RNN &  79.0  & -  & - \\
SA-CNN(This article) & 80.7 & 83.2 & \textbf{93.8} \\
\bottomrule
\hline
\end{tabular*}
\label{tab1}
\end{center}
\end{table}

As shown in the table.3, on the MR and CR datasets, the model presented in this paper outperformed other neural network structures, leading the authors to conclude that, due to a lack of computational resources, the parameter search range was restricted and the optimal network was not found. Nevertheless, the performance of the convolutional neural network was utilized, and on the TREC dataset, the model SA-CNN achieved the highest accuracy rate. Under the assumption of using the same model structure, the experimental results demonstrate that using the simulated annealing algorithm to find the optimal hyperparameters not only reduces the tediousness of manual parameter tuning, but also yields better parameters than manual tuning if the search range is correct, thereby achieving a high test accuracy.
\subsubsection{Discussion of experimental results}
In order to comprehend the characteristics of the ideal hyperparameters discovered by the simulated annealing technique, the following tables list the top 3 optimal hyperparameters sought by the algorithm in the TREC dataset.
\begin{table}[H]
\caption{TOP3 hyperparameters on TREC dataset.}
\begin{center}
\begin{tabular*}{0.7\linewidth}{c c c c}
\toprule
\hline
\textbf{Hyperparameters}& \textbf{Top1}& \textbf{Top2}& \textbf{Top3}\\
\hline
\midrule
filter num of win 3 & 100 & 100 & 32  \\
filter num of win 4 & 64 & 64 & 50  \\
filter num of win 5 & 32 & 64 & 50  \\
activation function & tanh & Relu & Relu.85  \\
Learning Rate & 0.002 & 0.002 & 0.002 \\
Dropout Rate & 0.4 & 0.1 & 0.4 \\
Batch size & 64 & 64 & 64 \\
\bottomrule
\hline
\end{tabular*}
\label{tab1}
\end{center}
\end{table}
Compared to manual tuning, the simulated annealing algorithm may search for hyperparameter combinations that one would not ordinarily consider; for instance, the number of CNN convolutional kernels for the Top1 model on the TREC dataset is 100, 64, and 32 for three different strings, which is a combination that one would not ordinarily consider. Therefore, by utilizing the simulated annealing process to optimize the neural network's hyperparameters, it is theoretically possible to acquire hyperparameter combinations that have not been considered or disregarded based on prior experience, and so achieve improved performance.

    \section{CONCLUSION}
In this article, we proposed a machine learning method combining simulated annealing and convolutional neural networks. The main goal is to adjust the hyperparameters of neural networks using simulated annealing in order to prevent manual tuning parameters into local optima, thus failing to enhance neural network performance. The experimental results demonstrate that the method of implementing simulated annealing to tune the hyperparameters of a neural network is effective in overcoming the constraints of manual parameter tuning, is practical, and can be theoretically applied to additional natural language processing problems. Due to the limited resource space and the different cooling rate for defining the initialization, which may result in different time costs and architecture, the final solution may only be an approximate optimal solution, and this approximation may differ. Consequently, the simulated annealing method may be integrated with other algorithms or the multi-objective simulated annealing algorithm may be further optimized, thereby further enhancing the efficiency of the simulated annealing algorithm on neural network optimization.

\bibliographystyle{unsrt}
\bibliography{reference.bib} 

\end{document}